\pdfoutput=1

\documentclass[11pt]{article}

\usepackage{graphicx}
\usepackage{booktabs}

\usepackage[preprint]{acl}

\usepackage{times}
\usepackage{latexsym}

\usepackage[T1]{fontenc}

\usepackage[utf8]{inputenc}

\usepackage{microtype}

\usepackage{inconsolata}
\usepackage{tabularx}
\newcolumntype{R}[1]{>{\raggedleft\arraybackslash}p{#1}}

\title{The \textit{Idola Tribus} of AI: Large Language Models tend to perceive order where none exists}

\author{Shin-nosuke Ishikawa \\
  Graduate School of Artificial Intelligence and Science, Rikkyo University\\
Strategic Digital Business Unit, Mamezou Co., Ltd.\\
  \texttt{shinnosuke-ishikawa@rikkyo.ac.jp} 
  \\\AND
  Masato Todo \and Taiki Ogihara\\
Strategic Digital Business Unit,\\ Mamezou Co., Ltd.\\\And
  Hirotsugu Ohba\\
Graduate School of Artificial Intelligence\\ and Science, Rikkyo University\\
}

\begin{document}
\maketitle
\begin{abstract}
We present a tendency of large language models (LLMs) to generate absurd patterns despite their clear inappropriateness in a simple task of identifying regularities in number series. Several approaches have been proposed to apply LLMs to complex real-world tasks, such as providing knowledge through retrieval-augmented generation and executing multi-step tasks using AI agent frameworks. However, these approaches rely on the logical consistency and self-coherence of LLMs, making it crucial to evaluate these aspects and consider potential countermeasures.
To identify cases where LLMs fail to maintain logical consistency, we conducted an experiment in which LLMs were asked to explain the patterns in various integer sequences, ranging from arithmetic sequences to randomly generated integer series. While the models successfully identified correct patterns in arithmetic and geometric sequences, they frequently over-recognized patterns that were inconsistent with the given numbers when analyzing randomly generated series. This issue was observed even in multi-step reasoning models, including OpenAI o3, o4-mini, and Google Gemini 2.5 Flash Preview Thinking.
This tendency to perceive non-existent patterns can be interpreted as the AI model equivalent of \textit{Idola Tribus} and highlights potential limitations in their capability for applied tasks requiring logical reasoning, even when employing chain-of-thought reasoning mechanisms.
\end{abstract}

\section{Introduction}
 Recent achievements of large language models (LLMs) have raised expectations that they can perform well across a wide range of tasks in human activities, reducing labor and duties through automation enabled by artificial intelligence (AI) technologies \citep{minaee2024, wan2024}.
In particular, the AI agent framework is a promising approach for handling complex tasks by integrating LLMs with external systems, enabling self-situation understanding and action planning \citep{xi2025, wooldridge1995}. To develop autonomous systems capable of executing complex real-world tasks, a deep understanding of LLM behavior is essential.

A key advantage of LLMs over other AI systems is that their input and output are in human-understandable natural language. Users can provide task instructions through everyday conversation without strict formatting, much like assigning tasks to colleagues. 
There is ongoing discussion that LLMs exhibit behavior that can only be interpreted as genuine language understanding, making it increasingly difficult to dismiss the possibility that they truly comprehend natural language \citep{mitchell2023}.
Additionally, results can be received through a conversational interface, allowing users to ask follow-up questions. 
In this context, it is crucial to ensure that LLMs accurately interpret human-provided prompts and generate reasonable and reliable outputs, rather than merely producing ``language-like'' lists of information. 
To assess this aspect, \citep{kong2024} proposed a method for aligning LLMs with user objectives to enhance human-LLM communication. Similarly, the \citet{lcm2024} introduced a new framework, the Large Concept Model, which aims to address the abstract nature of natural language communication by incorporating higher-level semantic representations called ``concepts.''

One widely recognized issue in LLM applications is hallucination, where LLM outputs contain untruthful information \citep{huang2025, zhang2023}. While one possible cause is the limited knowledge coverage in training data, studies have shown that hallucinations can occur even when an LLM possesses sufficient knowledge to generate correct answers \citep{simhi2024}.  
To address knowledge gaps, various frameworks have been proposed to enhance LLMs with external information through input prompts, such as in-context learning \citep{brown2020} and retrieval-augmented generation \citep{gao2024}.

Another critical issue with LLMs is ensuring a reasonable thinking process and self-consistency. Leaps in logic or unsupported conjectures can lead to false conclusions and may ultimately cause task execution failures. Chain-of-thought (CoT) prompting techniques have been proposed to address this issue by encouraging step-by-step reasoning \citep{wei2022, kojima2022}.  
Recently, major LLM developers have introduced "thinking" models, such as OpenAI o3, o4-mini \citep{o3o4mini}, and Google Gemini 2.5 Flash Preview Thinking \citep{gemini25flash}, which incorporate built-in multi-step self-evaluation and modification mechanisms based on the CoT concept.

A reasonable thinking process involves not only deduction but also induction, both of which are crucial for executing real-world tasks. In practical societal settings, strict procedural definitions for tasks are often limited, making it necessary to formulate hypotheses and proceed accordingly. Effective hypothesis formation requires not only consistency with the provided information but also the ability to abstract key patterns and principles. Without proper abstraction, misunderstandings or biased assumptions can critically impact task execution.

In this paper, we present an experiment to investigate the capabilities of LLMs, including thinking models, in forming hypotheses through precise information understanding and pattern abstraction. We use a simple task of identifying regularities in number series, allowing us to evaluate LLMs' pattern recognition ability independently of hallucinations related to knowledge accuracy.

Francis Bacon, in his famous work \textit{Novum Organum} \citep{bacon1620}, identified the tendency of biased belief and referred to it as \textit{Idola}, an inherent aspect of human nature. In particular, the tendency to over-recognize patterns in randomness is called \textit{Idola Tribus} (Idols of the Tribe). 
The motivation of this research is to examine whether \textit{Idola Tribus}, as described in the following citation, is a relevant concern for AI systems, particularly LLMs.

\begin{quote}\itshape
``The human understanding is of its own nature prone to suppose the existence of more order and regularity in the world than it finds. And though there be many things in nature which are singular and unmatched, yet it devises for them parallels and conjugates and relatives which do not exist.'' 
\par\hfill--- Francis Bacon, Novum Organum, Aphorism XLV. (1620) \footnote{Original Latin edition \cite{bacon1620}; English translation from
\cite{bacon190x}.}
\end{quote}

\section{Related Work}
\subsection{Evaluation and Enhancement of Logical Reasoning Capabilities}
Various studies have been conducted to evaluate logical consistency in natural language processing, and several datasets have been introduced to assess the logical reasoning capabilities of LLMs. ReClor  \citep{yu2020}, LogiQA \citep{liu2020}, and LogiQA2.0 \citep{liu2023} are representative datasets for evaluating self-consistent deductive reasoning. These datasets assess logical capabilities through predefined answer choices, where the correct answers are explicitly determined. While \citep{creswell2022} introduced inductive tasks to evaluate LLMs' logical reasoning capabilities, the correct answers in these tasks are still explicitly defined.

Our approach focuses on evaluating the inductive information abstraction capabilities of LLMs, making tasks without fixed answers more appropriate. In this context, we designed an experiment on regularity identification in number series, highlighting the contrast with previous studies.

In addition, several studies have aimed to enhance the logical reasoning abilities of LLMs. \citet{dalvi2022} proposed a framework for explaining the reasoning process using entailment trees, while \citet{pan2023} attempted to improve logical reasoning by integrating LLMs with symbolic solvers. These approaches could potentially contribute to enhancing the information abstraction capabilities evaluated in this paper.

\subsection{Mathematical Capability}
The task of identifying regularities in numerical series, as addressed in this paper, can also be interpreted as part of mathematical problem-solving. Various datasets, ranging from grade school to college-level mathematics, have been developed to assess LLMs' mathematical capabilities, including AQUA-RAT \citep{ling2017}, MATH \citep{hendrycks2021}, GSM8K \citep{cobbe2021gsm8k}, GSM-Plus \citep{li2024gsmplus}. Recently, it has been reported that an LLM-based system achieved a gold medalist performance in the Mathematical Olympiad \citep{chervonyi2025}.

Basic number series problems, such as arithmetic and geometric sequences, are included in the MATH dataset. However, these problems are designed to be solved using explicit rules and are fundamentally different from the open-ended task presented in this paper. While this study is related to mathematical capability, its primary objective is to evaluate self-consistency free from biases, rather than focusing solely on the mathematical aspect.
In addition, our task does not require the ability to convert real-world problem statements into mathematical expressions for computation.

\subsection{Biases in LLMs}
There are many indications of biases in LLMs, including gender \citep{wan2023}, political \citep{rozado2024}, and cultural \citep{tao2024} biases. 
Although this stems from a different perspective than the concept of \textit{Idola Tribus}, which is the main focus of this study, LLMs have also been reported to exhibit behaviors analogous to human cognitive biases \citep{echterhoff2024, shaikh2024}. Since LLMs are trained on texts written by humans, these biases can be seen as a negative legacy inherited from human data. The findings of this study should be considered as part of these biases and represent an issue that needs to be addressed.

\section{Method}
We conducted an experiment in which LLMs identified regularities in a prepared list of number series and then evaluated whether their descriptions accurately explained the series to investigate \textit{Idola Tribus}, over-recognition biases. The details of the regularity identification configuration and evaluation method are provided in the following subsections.

\subsection{Configuration for Regularity Identification in Number Sequences}
To evaluate how LLMs generalize provided information and identify patterns, as well as how they make hasty and inaccurate generalizations leading to false patterns, we prepared several categories of integer sequences, ranging from easily recognizable patterns to cases that are nearly impossible to define with a clear rule. Table~\ref{tab_dataset} presents these categories along with descriptions, the number of series prepared, and examples. We fixed several sets of number series for each category and had multiple LLMs perform identical tasks using the same numbers. The total number of numerical series tested was 724. In general, we used positive integers up to 100 as the values in the series, except for the geometric series categories, to simplify the experimental settings and clearly assess whether LLM responses exhibit bias.
\begin{table*}[h!]
\centering
\begin{tabular}{R{4mm}R{10mm}p{17mm}p{77mm}p{30mm}}
\toprule
\textbf{ \#} & \textbf{Count}& \textbf{Category} & \textbf{Description} & \textbf{Example} \\
\hline
1 &81& arithmetic & Arithmetic series with positive integer common differences ranging from 1 to 9. The number of patterns is 81, consisting of 9 common differences (1 to 9) multiplied by 9 first terms (1 to 9).
&$8, 9, 10, 11, 12, $... \\
2 &81& geometric & Geometric series with integer common ratios from $-5$ to 5, excluding 0 and 1.  The number of patterns is 81, consisting of 9 common ratio ($-5$ to $-1$ and 2 to 5) multiplied by 9 first terms (1 to 9).&$3, -6, 12, -24, 48, $... \\
3 &100& difference & Number series in which the differences between consecutive terms form arithmetic sequences with positive integer common differences ranging from 1 to 9.&$4, 7, 11, 16, 22, $... \\
4 &81& quasi-arithmetic &  Almost arithmetic series with the same conditions as \#1, but with one term deviating by 1 from the expected pattern. & $8, 10, 10, 11, 12,$ ...\\
5&81& quasi-geometric & Almost geometric series with the same conditions as \#2, but with one term deviating by 1 from the expected pattern. & $3,-6,11, -24, 48, $...\\
6 &100& quasi-difference & Almost difference series with the same conditions as \#3, but with one term deviating by 1 from the expected pattern. & $4, 8, 11, 16, 22, $...\\
7 &100& random-increasing & Randomly generated increasing integer sequences. Not applicable to \#1--\#6.& $17, 25, 33, 43, 50, $...\\
8 &100& random & Randomly generated integer sequences. Not applicable to \#1--\#7.& $54, 1, 78, 7, 49, $...\\
\hline
\textbf{Total} &724&&&\\
\bottomrule
\end{tabular}
\caption{List of number series categories for the regularity identification test, ranging from easily identifiable arithmetic series to random series with no clear order.}
\label{tab_dataset}
\end{table*}

For the easier cases, we prepared arithmetic, geometric, and difference series with difference sequences. The arithmetic series were generated with first terms and common differences selected randomly from integers between 1 and 9. In the geometric series, the common ratio was chosen from integers between  $-5$ and 5, excluding -$1$, 0, and 1, to prevent excessively large absolute values in the sequence. The first term was randomly selected from integers between 1 and 9, following the same setting as the arithmetic series. For the difference series, the first term, along with both the first term and the common difference of the difference sequence, were randomly selected from integers between 1 and 9.
We also prepared quasi-ordered cases---arithmetic, geometric, and difference series with a single-term error of +1 or -1---to investigate whether LLMs recognize these errors and distinguish them from purely ordered series.

In addition to the number series categories described above, we prepared two categories: random-increasing and random. The random-increasing series consist of randomly generated numbers with the condition that each term is greater than the previous one, while the random series have no such constraint. These series generally do not exhibit clear regularity and are used to assess whether LLMs incorrectly perceive false patterns.
To ensure distinct categorization, we designed these series to avoid overlap with other categories (e.g., random-increasing series are not arithmetic, geometric, or difference series, and random series do not belong to the random-increasing category).
The random-increasing series are generated by adding random integers between 1 and 10 to the previous term, starting from an initial term randomly selected between 2 and 18. The random series are generated using random integers between 1 and 99.

We selected the latest high-performance LLMs, widely used across various applications, for evaluation: Open AI GPT-4.1 \citep{gpt4.1}, o3, o4-mini \citep{o3o4mini} and Google Gemini 2.5 Flash  Preview Thinking \citep{gemini2023, gemini25flash} (abbreviated as ``Gemini2.5'' in the tables).
The o3, o4-mini, and Gemini 2.5 Flash Preview Thinking models incorporate the latest multi-step CoT reasoning techniques to enhance logical consistency. 
Additionally, we included Llama 3.3 \citep{llama} as a representative of high-performance open models.
The versions used were gpt-4.1-2025-04-14 for GPT-4.1, o3-2025-04-16 for o3, o4-mini-2025-04-16 for o4-mini, and gemini-2.5-flash-preview-04-17-thinking for Gemini 2.5 Flash Preview Thinking.

\begin{figure}[h!]
\centering
\includegraphics[width=7.8cm]{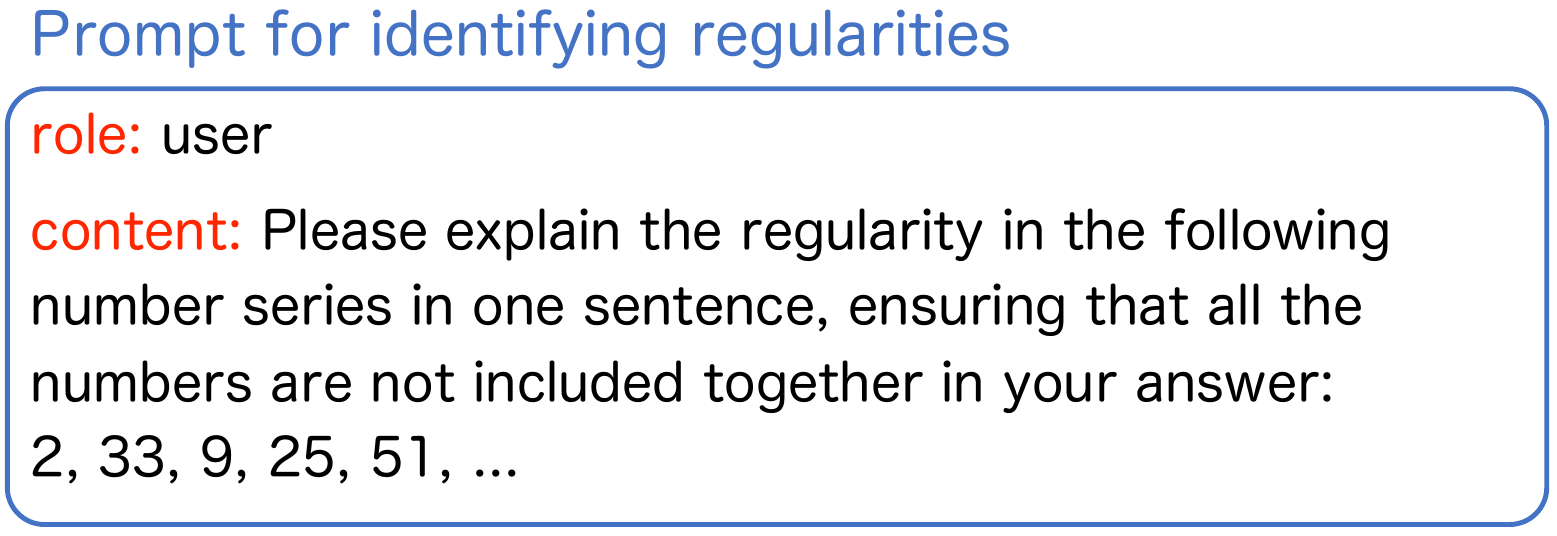}
\caption{Prompt for the regularity identification task, shown for a sample case of a random series: 2, 33, 9, 25, 51, ... }
\label{fig_prompt1a}
\end{figure}

Figure~\ref{fig_prompt1a} shows an input prompt with a simple instruction for the regularity identification task. Each prompt included five values from a given series. Regularity becomes easier to explain when only four or fewer values are provided, such as when fitting a third-order polynomial function, while randomness becomes clearer when six or more values are provided in the random series. The same prompt was used across all models. The output is restricted to a single sentence via the prompt for ease of evaluation.
We standardized the approach by providing instructions solely through user prompts, without using system prompts.

\subsection{Evaluation Method for Inferred Regularities}
As the next step, we evaluated the regularity descriptions obtained using the method described in the previous subsection to determine how many number series were successfully generalized and abstractly described correctly. While the validated cases confirm that the LLMs performed well, analyzing the failed cases is crucial for understanding their tendencies.  
Therefore, the evaluation has two main objectives: assessing regularity identification capabilities and analyzing the patterns in failure cases.

We emphasize that there is no perfect autonomous evaluation method with flawless accuracy for determining the validity of regularity descriptions. However, maintaining quality and consistency across 724 $\times$ 5 $=$ 3,620 regularity descriptions through human evaluation alone is challenging in terms of reproducibility and consistency. Therefore, we chose to use LLMs as evaluators to ensure the experiment remains reproducible and consistent, even if their accuracy is limited.
The concept of using LLMs as evaluators, known as ``LLM-as-a-Judge,'' has become increasingly common \citep{gu2025}. \citet{zheng2023} demonstrated that the agreement between LLM-as-a-Judge and human annotators is comparable to inter-annotator agreement, indicating that LLMs have the capability to evaluate LLM-generated descriptions.

\begin{figure}[h!]
\centering
\includegraphics[width=7.8cm]{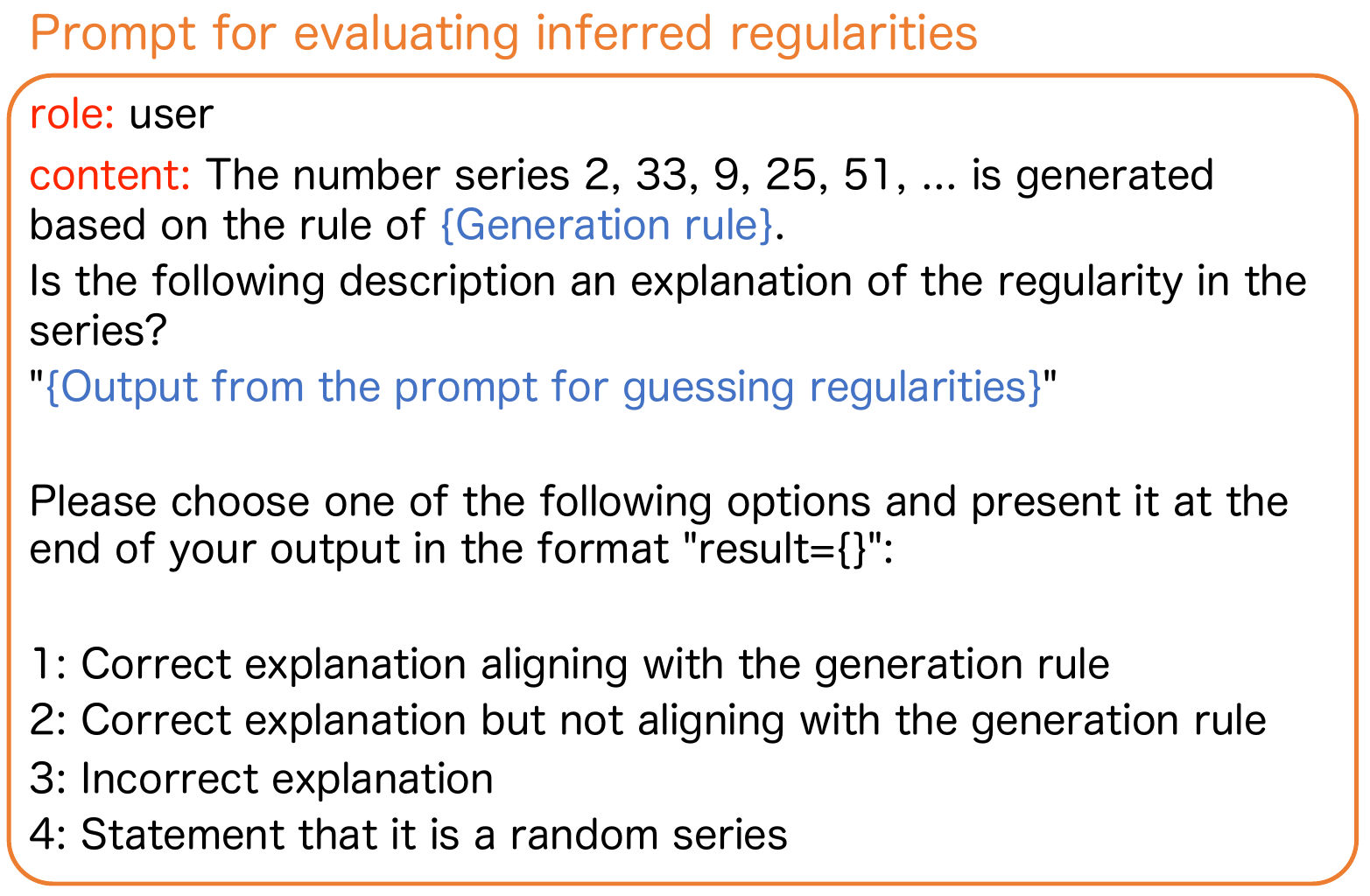}
\caption{Prompt for regularity description evaluation, shown for a sample case of a random series: 2, 33, 9, 25, 51, ... corresponding to the case in Fig.~\ref{fig_prompt1a}.}
\label{fig_prompt2}
\end{figure}

The prompt for LLM evaluation of the regularity descriptions is shown in Fig.~\ref{fig_prompt2}. 
We designed the evaluation prompt not only to assess the validity of regularity descriptions but also to analyze their characteristics, tendencies, and the potential to avoid invalid outputs. To support this, we distinguished between valid descriptions that align with the preset category and those that do not. Additionally, we included an evaluation option for descriptions that state the series is random. Since the dataset includes a random series category, models are not required to invent plausible regularities for these cases. We analyzed how often the outputs correctly identify such series as random.
In summary, we defined four evaluation options (Fig.~\ref{fig_prompt2}): (1) correct explanation aligning with the preset category, (2) correct explanation not aligning with the preset category, (3) incorrect explanation, and (4) statement that the series is random.

We calculated the success rate based on the evaluation results to assess how well the target LLMs can generalize the provided numerical information and describe the regularities.
In this calculation, outputs corresponding to options 1 and 2 are both considered successful explanations, regardless of whether they align with the preset category.
For categories other than arithmetic, geometric, and difference series, we also treat a statement that the series is random (option 4) as a successful explanation, as these categories do not have clearly describable regularities.

For the evaluation models, we used the o3 model. This selection was based on a preliminary experiment using author-annotated descriptions, where all candidate models performed adequately, but o3 showed the best performance within the test setting. Details of the preliminary experiment are provided in Appendix~\ref{sec_preexp}.

\section{Result}
We successfully obtained 3,620 regularity descriptions for 724 number series shown in Table~\ref{tab_dataset} using the five regularity identification LLMs, along with evaluation results for all descriptions. A summary of the success rates, aggregated based on the evaluation results, is presented in Table~\ref{tab_rate}.  
\begin{table*}[h!]
\centering
\begin{tabular}{lrrrrrr}
\toprule
\textbf{Category} & \multicolumn{1}{c}{\textbf{GPT-4.1}} & \multicolumn{1}{c}{\textbf{o3}} & \textbf{o4-mini} & \multicolumn{1}{c}{\textbf{Gemini2.5}}&\multicolumn{1}{c}{\textbf{Llama3.3}} &\multicolumn{1}{c}{\textbf{Average}} \\
\hline
arithmetic&\textbf{100\%}&\textbf{100\%}&\textbf{100\%}&\textbf{100\%}&\textbf{100\%}&100\%\\
geometric&\textbf{100\%}&98.8\%&\textbf{100\%}&\textbf{100\%}&95.1\%&98.8\%\\
difference&36.0\%&99.0\%&99.0\%&\textbf{100\%}&93.0\%&85.4\%\\
quasi-arithmetic&16.0\%&51.9\%&56.8\%&\textbf{63.0\%}&21.0\%&41.7\%\\
quasi-geometric&11.1\%&55.6\%&25.9\%&\textbf{65.4\%}&12.3\%&34.1\%\\
quasi-difference&4.0\%&64.0\%&64.0\%&\textbf{70.0\%}&7.0\%&41.8\%\\
random-increasing&7.0\%&\textbf{66.0\%}&56.0\%&44.0\%&4.0\%&35.4\%\\
random&8.0\%&\textbf{52.0\%}&43.0\%&31.0\%&0.0\%&26.8\%\\
\hline
\textbf{Total} &33.0\%&\textbf{73.1\%}&67.8\%&70.6\%&39.9\%&56.9\%\\
\bottomrule
\end{tabular}
\caption{Success rates of regularity identification for each category and LLM, based on the described evaluation method. Boldface indicates the highest success rate achieved by an individual model for each category.}
\label{tab_rate}
\end{table*}

All five models correctly identified the regularities in all arithmetic series, achieving a 100\% success rate, and performance was also high for the geometric series. For the difference series, o3, o4-mini, Gemini 2.5 Flash Preview Thinking and Llama 3.3 maintained strong performance, while the success rate declined for GPT-4.1.  
For the quasi-arithmetic, quasi-geometric, and quasi-difference series, the success rate decreased across all LLMs, reflecting the fact that these series do not exhibit clear regularities. In particular, the non-thinking models GPT-4.1 and Llama 3.3 performed worse than the self-iterative reasoning models o3, o4-mini, and Gemini 2.5 Flash Preview Thinking.  
In the random-increasing and random categories, the trend of thinking models outperforming non-thinking models remained consistent with the quasi-ordered series cases.
Overall, the success rate was high for the thinking models, with comparable performance among them.
However, the success rate for the random series remained notably low across all LLMs.

The fact that LLMs provide incorrect explanations for random series clearly demonstrates their tendency to overestimate regularities when interpreting information---specifically, the numerical values in the series used in the experiment---indicating the presence of \textit{Idola Tribus} in LLMs.
As a result, our findings reveal a tendency to perceive ordered patterns that are inconsistent with the provided information---clear evidence that these AI models exhibit \textit{Idola Tribus}.

\begin{figure}[h!]
\centering
\includegraphics[width=7.8cm]{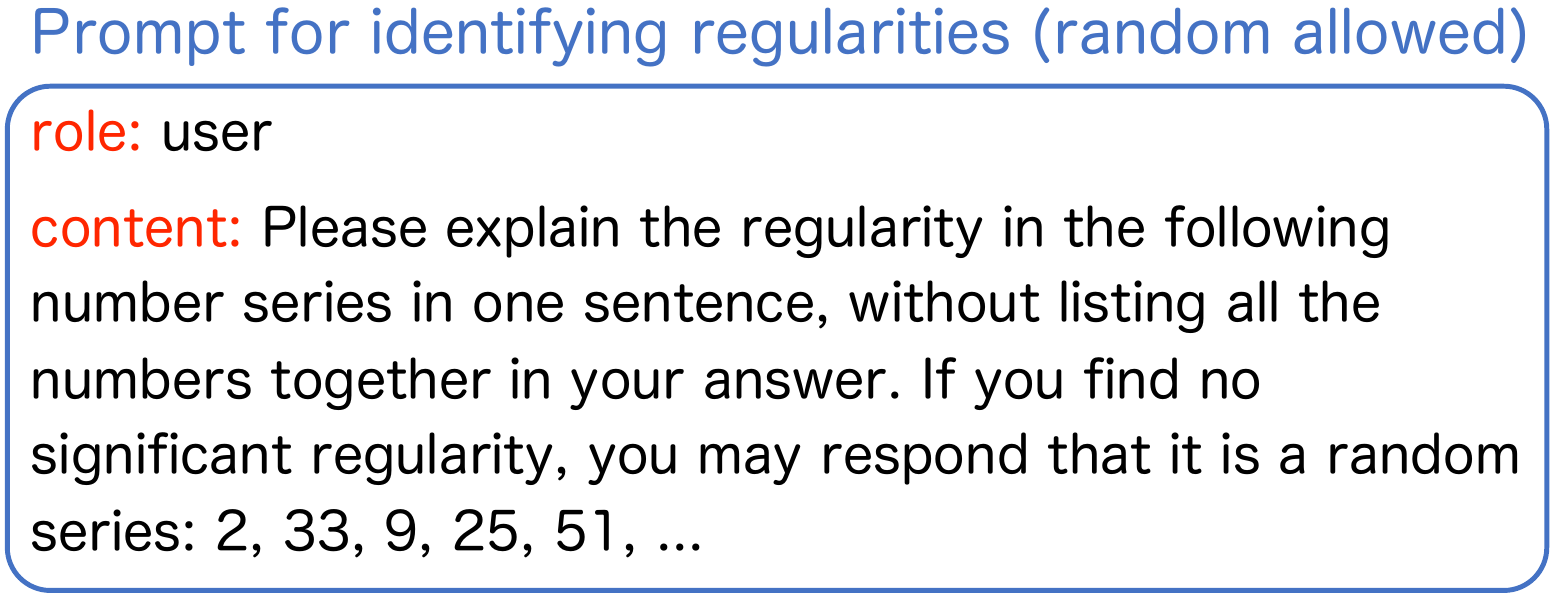}
\caption{Prompt for the regularity identification task, shown for a sample case of a random series: 2, 33, 9, 25, 51, ... }
\label{fig_prompt1b}
\end{figure}

To investigate whether LLMs tend to force a pattern even when they fail to find a plausible one, we conducted an additional experiment by modifying the regularity identification prompt to explicitly allow the series to be random (Fig.~\ref{fig_prompt1b}). 
Table~\ref{tab_rate_random} shows a comparison of the rate at which each model explains that the series is random, using the original prompt (Fig.~\ref{fig_prompt1a}) and the random-allowing prompt (Fig.\ref{fig_prompt1b}).
We observe that the rate of random-series explanations increases significantly with GPT-4.1, o3, o4-mini, and LLaMA 3.3 in categories other than arithmetic, geometric, or difference series.
This suggests that prompting with the option to declare a series as random---indicating no significant regularity---encourages these models to state more confidently when they find no clear pattern.
In contrast, Gemini 2.5 Flash Preview Thinking does not show a similar change.
\begin{table*}[h!]
\centering
\begin{tabular}{lrrrrrr}
\toprule
\textbf{Category} & \multicolumn{1}{c}{\textbf{GPT-4.1}} & \multicolumn{1}{c}{\textbf{o3}} & \textbf{o4-mini} & \multicolumn{1}{c}{\textbf{Gemini2.5}}&\multicolumn{1}{c}{\textbf{Llama3.3}} &\multicolumn{1}{c}{\textbf{Average}} \\
\hline
quasi-arithmetic&0.0\%&0.0\%&0.0\%&0.0\%&0.0\%&0.0\%\\
quasi-geometric&0.0\%&0.0\%&0.0\%&0.0\%&0.0\%&0.0\%\\
quasi-difference&0.0\%&0.0\%&0.0\%&0.0\%&0.0\%&0.0\%\\
random-increasing&1.0\%&\textbf{9.0\%}&6.0\%&8.0\%&0.0\%&4.8\%\\
random&5.0\%&\textbf{28.0\%}&14.0\%&17.0\%&0.0\%&12.8\%\\
\hline
\textbf{Total} &1.3\%&\textbf{8.0\%}&4.3\%&5.4\%&0.0\%&3.8\%\\
\bottomrule
\end{tabular}

\medskip

\begin{tabular}{lrrrrrr}
\toprule
\textbf{Category} & \multicolumn{1}{c}{\textbf{GPT-4.1}} & \multicolumn{1}{c}{\textbf{o3}} & \textbf{o4-mini} & \multicolumn{1}{c}{\textbf{Gemini2.5}}&\multicolumn{1}{c}{\textbf{Llama3.3}} &\multicolumn{1}{c}{\textbf{Average}} \\
\hline
quasi-arithmetic&18.5\%&13.6\%&\textbf{30.9\%}&0.0\%&2.5\%&13.1\%\\
quasi-geometric&\textbf{13.6\%}&3.7\%&9.9\%&0.0\%&6.2\%&6.7\%\\
quasi-difference&6.0\%&33.0\%&\textbf{59.0\%}&0.0\%&0.0\%&19.6\%\\
random-increasing&38.0\%&56.0\%&\textbf{66.0\%}&5.0\%&17.0\%&36.4\%\\
random&83.0\%&87.0\%&\textbf{93.0\%}&30.0\%&86.0\%&75.8\%\\
\hline
\textbf{Total} &33.1\%&41.1\%&\textbf{54.3\%}&7.6\%&23.8\%&32.0\%\\
\bottomrule
\end{tabular}
\caption{Rate of explanations stating that the target is a random series (evaluation result option 4). Upper panel: results with the original prompt (Fig.~\ref{fig_prompt1a}); lower panel: results with the random-allowing prompt (Fig.~\ref{fig_prompt1b}).
Boldface indicates the highest rate by an individual model for each category.}
\label{tab_rate_random}
\end{table*}

The success rates with the random-allowing prompt configuration are shown in Table~\ref{tab_rate_p2}.
Compared to Table~\ref{tab_rate}, the improvements correspond to the increase in random-series explanations reported in Table~\ref{tab_rate_random}.
The o3 and o4-mini models showed comparable performance, achieving the highest success rates among the models.
Nevertheless, success rates for quasi-ordered series remain significantly lower than those for their corresponding ordered cases across all models.
Gemini 2.5 Flash Preview Thinking did not show a notable improvement, as the number of random-series explanations did not increase substantially.
\begin{table*}[h!]
\centering
\begin{tabular}{lrrrrrr}
\toprule
\textbf{Category} & \multicolumn{1}{c}{\textbf{GPT-4.1}} & \multicolumn{1}{c}{\textbf{o3}} & \textbf{o4-mini} & \multicolumn{1}{c}{\textbf{Gemini2.5}}&\multicolumn{1}{c}{\textbf{Llama3.3}} &\multicolumn{1}{c}{\textbf{Average}} \\
\hline
arithmetic&\textbf{100\%}&\textbf{100\%}&\textbf{100\%}&\textbf{100\%}&\textbf{100\%}&100\%\\
geometric&\textbf{100\%}&\textbf{100\%}&98.8\%&\textbf{100\%}&\textbf{100\%}&99.8\%\\
difference&45.0\%&\textbf{100\%}&\textbf{100\%}&\textbf{100\%}&97.0\%&88.4\%\\
quasi-arithmetic&40.7\%&66.7\%&\textbf{82.7\%}&54.3\%&24.7\%&53.8\%\\
quasi-geometric&24.7\%&\textbf{69.1\%}&45.7\%&66.7\%&17.3\%&44.7\%\\
quasi-difference&9.0\%&84.0\%&\textbf{86.0\%}&69.0\%&11.0\%&51.8\%\\
random-increasing&46.0\%&85.0\%&\textbf{91.0\%}&40.0\%&20.0\%&56.4\%\\
random&83.0\%&94.0\%&\textbf{97.0\%}&30.0\%&86.0\%&78.0\%\\
\hline
\textbf{Total} &55.0\%&87.7\%&\textbf{88.3\%}&68.9\%&56.6\%&71.3\%\\
\bottomrule
\end{tabular}
\caption{Success rates of regularity identification for each category and LLM using the random-allowing prompt (Fig.~\ref{fig_prompt1b}).
Boldface indicates the highest success rate achieved by an individual model for each category.}
\label{tab_rate_p2}
\end{table*}

\section{Discussion}
As shown in the previous section, we presented a clear and comprehensive analysis---demonstrating for the first time---that LLMs occasionally generate false patterns that contradict the information they are given.
When a clear regularity cannot be identified, the model should explicitly state this, rather than providing potentially inaccurate explanations.
The lack of random-series explanations under the original prompt clearly illustrates this issue, where models incorrectly assert the presence of a pattern instead of acknowledging the absence of an identifiable structure.

One notable finding is that the tendency to over-recognize patterns also appeared in the thinking models, despite their higher success rates compared to non-thinking models (Table~\ref{tab_rate}). This suggests either that the thinking models lack sufficient capability to verify regularity hypotheses or that they assert false regularities even when they recognize them as incorrect.  
This question can be examined using the o3 results, as o3 was used for both regularity identification and evaluation. We found that o3 judged its own identified regularities as valid in only 52 out of 100 cases within the random category. This indicates that the o3 model generated false patterns even in cases where it itself recognized them as incorrect.
The failure to reflect self-evaluation results may be related to fundamental differences between the CoT process and human thinking \citep{bao2024}.

By comparing correct regularity explanations that align and do not align with the preset categories (evaluation result options 1 and 2), we found that in the arithmetic, geometric, and difference series, almost all correct explanations aligned with the preset categories. Since the regularities in these categories are simple and well-defined, it is reasonable that LLMs did not attempt to explain them using alternative patterns.
In contrast, for the quasi-ordered categories, which include one deviation from a clear regularity, there were comparable numbers of correct explanations both aligning and not aligning with the preset categories.
This suggests that the evaluated models demonstrate strong performance in their ability to identify valid regularities, even when those regularities were not explicitly intended in the series generation process.

For the o3 and o4-mini models, a greater variety of highly creative ideas were generated compared to other models, including interpretations based on atomic numbers, football players, piano, tarot, the Holy Bible, and telephone country codes---even though these were inconsistent with the given series. This level of creativity could be a significant advantage if the model’s internal mechanism for logical self-validation functioned reliably.
At present, however, both o3 and o4-mini tend to validate ideas they are not well-equipped to evaluate, leading to misinterpretations and false confirmations. It has been reported that LLMs often struggle with tasks such as recognizing characters in words, counting words, or identifying letter sequences\citep{qin2023}---weaknesses that may contribute to the observed failures. In such cases, the models' wide-ranging ideas ultimately lead to lower-quality outputs.

Based on the behavioral changes observed with the random-allowing prompt, it is plausible that this bias stems from an implicit compulsion in LLMs to always provide an answer in response to a given instruction. This tendency can lead to false outputs resembling confabulation or improvisation---similar to hallucinations.
In this context, frameworks designed to mitigate hallucinations may offer useful insights.
Several studies have explored such approaches, including prompting LLMs to state explicitly when they do not have certain knowledge \citep{zhang2024idontknow}, or even to explain why they cannot answer a given question \citep{deng2024}. 
If LLMs were able to state that they do not know the regularities in a number series---or to explain why they cannot identify them---the impact of this phenomenon could be significantly reduced.
Further investigation is needed to determine whether these concepts can be applied to cases of logical reasoning, rather than purely knowledge-based tasks.

The fact that the models with the best performance under the random-allowing prompt, o3 and o4-mini, achieve higher success rates for random series than for quasi-ordered series is notable.
Quasi-ordered series may appear to contain a regularity, as the deviation involves only a single value in the series. There are cases in which LLMs persist in trying to explain a pattern rather than recognizing the absence of a clear rule.
This tendency provides evidence that the behavior aligns with the characteristics of \textit{Idola Tribus}, suggesting that the phenomenon observed in LLMs shares the same nature.
Although the number of target LLMs is limited to five, we observed this tendency across all five models. This aspect also aligns with the concept of \textit{Idola Tribus}, which refers to shared tendencies among humans, suggesting a comparable tendency within the ``tribe'' of LLMs.

One possible cause of this phenomenon might be a tendency to process information efficiently during training or instruction tuning.
Providing additional prompts to help LLMs recognize this bias may reduce over-interpretation, although it could also limit their willingness to engage with more complex reasoning tasks.

To overcome this bias through approaches other than fine prompt adjustments, fine-tuning with strategies proposed in previous studies for enhancing logical reasoning capabilities could also be effective in addressing the pattern recognition biases identified in this paper \citep{zelikman2022, morishita2023}. Ideally, developing a bias-free model through fine-tuning would be preferable, as relying on a biased model requires constant vigilance to mitigate bias through prompting.  Although the fine-tuning configurations discussed in these studies are primarily designed to improve deductive reasoning, the core principle of maintaining logical consistency with all available information is equally important for addressing inductive biases. It has also been noted that model performance depends not only on the quantity of training data but, more importantly, on its quality \citep{ye2025}. In this context, exploring effective fine-tuning strategies is a key next step in tackling the issue.

\section{Conclusion}
We conducted an experiment in which LLMs identified regularities in various types of number series, including randomly generated ones, and discovered a tendency for LLMs to over-recognize patterns that do not fully explain the provided information. This tendency can be regarded as the LLM equivalent of \textit{Idola Tribus} in humans.  
LLMs tend to force themselves to explain patterns even when they do not find a plausible one, unless explicitly instructed to acknowledge the absence of regularities.
This tendency has been shown to share the same characteristic found in \textit{Idola Tribus} in humans, particularly in its greater likelihood to appear when the series seems ordered.

Future research is expected to focus on mitigating the impact of this bias, and further model development will be necessary to address it. 
Until this tendency is properly controlled, it is essential to remain aware of the issue when applying LLMs to practical tasks, just as humans must be cautious to avoid biases in general, in order to prevent unintended errors.

\section*{Limitation}
In this paper, we conducted an evaluation experiment on only five major LLMs to confirm the tendency for false pattern recognition. However, this does not guarantee that the same tendency exists in all current and future LLMs.  

We have presented results using only two variations for regularity identification, as shown in Figs.~\ref{fig_prompt1a} and \ref{fig_prompt1b}. Improved results may be achievable with more optimized prompts. Our intention was not to suggest that such tendencies are unavoidable, but rather to emphasize that users should be cautious of these incorrect recognitions.

Additionally, the evaluation method using LLMs selected in this study is not perfectly precise. As a result, the success rate values presented in this paper may not be highly accurate in absolute terms, though the overall trend and main findings remain unchanged.

\section*{Ethics Statement}
We investigated the tendencies of selected large language models and did not handle any personal information. This research focuses solely on logical biases, excluding considerations of gender, political, or cultural biases. Therefore, it does not raise any ethical concerns.

\section*{Acknowledgements}
We thank Eiiti Hanyuda for his suggestion and idea to assess the logical reasoning ability of LLMs using number series.
We also thank Masao Yasui for valuable discussions on the application of LLMs to real-world tasks.
This work was supported by JSPS KAKENHI Grant Number 24K15077. 
We utilized ChatGPT as an assistant to edit the text, aiming to improve the English expressions to make them more appropriate.

\bibliography{ishikawa2025}

\appendix
\section{Preliminary Experiment for Evaluation Method Investigation}\label{sec_preexp}
We conducted an experiment using four number series to investigate the evaluation method. Five LLMs from the main experiment were tasked with identifying regularities, and the validity of their descriptions was annotated by the authors as ground truth. The series were selected from the random, difference, and quasi-arithmetic categories, as descriptions for random and quasi-regular series are more challenging to evaluate. These four test cases were excluded from the main experiment.

Next, we tested the evaluation prompt on 21 selected descriptions, excluding vague or overly similar ones and including some results from the random-allowing prompt (Fig.~\ref{fig_prompt1b}), using the five models as candidate LLM-as-a-judge evaluators. Table~\ref{tab_preexp} the 21 descriptions used for testing. The expected evaluation results were annotated by the authors and are listed in Table~\ref{tab_preexp}. We note that it is not necessary to distinguish between annotations 1 and 4 for random series, as a correct description based on the generation rule (option 1) and a statement that the series is random (option 4) are effectively equivalent.
\begin{table*}[h!]
\small
\centering
\begin{tabular}{p{4mm}p{11mm}p{21mm}p{13mm}p{78mm}p{9mm}}
\toprule
\textbf{Exp ID} &\textbf{Category}&\textbf{Number series}& \textbf{Regularity identifier LLM}& \textbf{Regularity description} & \textbf{Author annotation}\\
\hline
P1&random&$2, 33, 9, 25, 51, $...&GPT-4.1&
The sequence alternates between adding 31 and subtracting 24 to generate the next number.
&\multicolumn{1}{r}{3}\\
P2&random&$2, 33, 9, 25, 51, $...&o4-mini&
After the initial 2, the sequence alternates between multiples of three whose other factor is a successive odd prime ($33 = 3\times11$, $51 = 3\times17$, …) and perfect squares of consecutive odd integers ($9 = 3^2$, $25 = 5^2$, …).
&\multicolumn{1}{r}{3}\\
P3&random&$2, 33, 9, 25, 51, $...&o3&
Treat the list as two interlaced sequences --- one made by repeatedly adding successive even squares ( $+4$, $+16$, $+36$, $+64$, … ) starting from 2, and the other by successively subtracting the corresponding odd squares ( $-9$, $-25$, $-49$, … ); when these two strands are woven together they yield the given numbers in the stated order.
&\multicolumn{1}{r}{3}\\
P4&random&$2, 33, 9, 25, 51, $...&Gemini2.5&
The sequence starts with 2 and 33; for n greater than two, term n equals the square of (2n-3), adding two if n is odd and five or greater.
&\multicolumn{1}{r}{2}\\
P5&random&$2, 33, 9, 25, 51, $...&o3&
No clear arithmetic, geometric, digit-manipulation or indexing rule links each element to its successor, so the list should be treated as a random assortment of numbers.
&\multicolumn{1}{c}{1 or 4}\\
P6&random&$2, 33, 9, 25, 51, $...&Gemini2.5&
There is no significant regularity in this series.
&\multicolumn{1}{c}{1 or 4}\\
P7&random&$85, 1, 3, 34, 1, $...&GPT-4.1&
The series alternates between subtracting 84 and then multiplying the result by the next odd number in sequence (1, 3, 5,…).
&\multicolumn{1}{r}{3}\\
P8&random&$85, 1, 3, 34, 1,$...&o3&
Begin with a two-digit number; write the GCD of its two digits, write the absolute difference of those digits, then make the next two-digit member by using that difference as the tens digit and the digital root of the digits’ sum as the units digit, and repeat.
&\multicolumn{1}{r}{2}\\
P9&random&$85, 1, 3, 34, 1,$...&o4-mini&
Each term after the initial two-digit number is generated by alternately summing its digits (to produce two separate one-digit entries) and then taking the two most recent one-digit entries to form the next two-digit term by writing their product followed by their sum.
&\multicolumn{1}{r}{3}\\
P10&random&$85, 1, 3, 34, 1,$...&Gemini2.5&
The series begins with eighty-five, and subsequent terms cycle through the values obtained by taking Fibonacci numbers at index one, four, and nine.
&\multicolumn{1}{r}{2}\\
P11&random&$85, 1, 3, 34, 1,$...&Llama3.3&
The series appears to alternate between a large number and the number 1, with the large numbers increasing in a specific pattern, while the small number remains constant at 1.
&\multicolumn{1}{r}{3}\\
P12&random&$85, 1, 3, 34, 1,$...&Llama3.3&
The series appears to be random, with no discernible pattern or regularity, as the numbers seem to be unrelated and do not follow a consistent sequence or mathematical operation.
&\multicolumn{1}{c}{1 or 4}\\

P13&quasi-arithmetic&$4, 13, 22, 32, 40,$...&GPT-4.1&
The pattern in the series is that the numbers increase alternately by 9 and then by 8.
&\multicolumn{1}{r}{3}\\
P14&quasi-arithmetic&$4, 13, 22, 32, 40,$...&o3&
Beginning with 4, each term is obtained by adding 9 to the previous one, so every member of the sequence is a number that leaves a remainder of 4 when divided by 9.
&\multicolumn{1}{r}{3}\\
P15&quasi-arithmetic&$4, 13, 22, 32, 40,$...&o4-mini&
Each consecutive gap cycles through adding nine, then nine again, then ten, and then eight.
&\multicolumn{1}{r}{2}\\
P16&quasi-arithmetic&$4, 13, 22, 32, 40,$...&Gemini2.5&
Each term after the first is found by adding an amount from a repeating four-step cycle.
&\multicolumn{1}{r}{2}\\

P17&difference&$5, 14, 29, 50, 77,$...&GPT-4.1&
In this series, each term increases by consecutive odd numbers multiplied by three: specifically, the difference between each consecutive pair rises by 6 each time (i.e., 9, 15, 21, 27, ...), so each term adds six more than the previous difference.
&\multicolumn{1}{r}{1}\\
P18&difference&$5, 14, 29, 50, 77,$...&o3&
The difference between consecutive terms rises by six each time (9, 15, 21, 27, …), so the nth term is given by $3n^2 + 2$.
&\multicolumn{1}{r}{1}\\
P19&difference&$5, 14, 29, 50, 77,$...&o4-mini&
Each term equals three times the square of its position in the sequence plus two.
&\multicolumn{1}{r}{2}\\
P20&difference&$5, 14, 29, 50, 77,$...&Gemini2.5&
The regularity is that the difference between consecutive terms increases by a fixed value.
&\multicolumn{1}{r}{1}\\
P21&difference&$5, 14, 29, 50, 77,$...&Llama3.3&
The series appears to be formed by adding 9, then 15, then 21, and so on, increasing the increment by 6 each time, which suggests a quadratic or second-order relationship between the terms.
&\multicolumn{1}{r}{1}\\
\bottomrule
\end{tabular}
\caption{Descriptions to be tested for the regularity description evaluator model, with author-provided annotations. The annotated numbers correspond to the options shown in Fig.~\ref{fig_prompt2}. The descriptions in P5, P6, and P12 were obtained using the random-allowing prompt  (Fig.~\ref{fig_prompt1b}).}
\label{tab_preexp}
\end{table*}

Table~\ref{tab_preexp_result} presents the results of the preliminary experiment. 
We found that all candidate models performed as evaluators with a certain level of accuracy, with o3 achieving the best performance. For o3, if we treat options 1 and 2 as equivalent---since both represent acceptable cases---there was only one error (Exp. P10), involving a discrepancy between acceptable and unacceptable judgments compared to the annotation. This was the fewest among the five candidates; therefore, we selected o3 as the LLM-as-a-judge evaluator for the main experiment.
\begin{table*}[h!]
\centering
\begin{tabular}{lrrrrrr}
\toprule
\textbf{Exp ID}  &\multicolumn{1}{c}{\textbf{GPT-4.1}} &\multicolumn{1}{c}{\textbf{o3}} & \multicolumn{1}{c}{\textbf{o4-mini}} &
\multicolumn{1}{c}{\textbf{Gemini2.5}} &\multicolumn{1}{c}{\textbf{Llama3.3}} & \multicolumn{1}{c}{\textbf{Annotation}} \\
\hline
P1&\textbf{3}&\textbf{3}&\textbf{3}&\textbf{3}&\textbf{3}&3\\
P2&1&\textbf{3}&4&2&1&3\\
P3&\textbf{3}&\textbf{3}&4&\textbf{3}&\textbf{3}&3\\
P4&1&1&4&1&3&2\\
P5&\textbf{4}&\textbf{4}&\textbf{1}&\textbf{1}&2&\multicolumn{1}{c}{1 or 4}\\
P6&\textbf{1}&\textbf{4}&\textbf{1}&\textbf{1}&\textbf{1}&\multicolumn{1}{c}{1 or 4}\\
P7&\textbf{3}&\textbf{3}&\textbf{3}&\textbf{3}&\textbf{3}&3\\
P8&\textbf{2}&1&1&1&\textbf{2}&2\\
P9&\textbf{3}&\textbf{3}&\textbf{3}&\textbf{3}&\textbf{3}&3\\
P10&1&3&3&\textbf{2}&3&2\\
P11&\textbf{3}&\textbf{3}&\textbf{3}&\textbf{3}&\textbf{3}&3\\
P12&\textbf{4}&\textbf{4}&\textbf{1}&\textbf{1}&\textbf{1}&\multicolumn{1}{c}{1 or 4}\\
P13&\textbf{3}&\textbf{3}&\textbf{3}&\textbf{3}&\textbf{3}&3\\
P14&2&\textbf{3}&\textbf{3}&\textbf{3}&\textbf{3}&3\\
P15&1&\textbf{2}&\textbf{2}&1&1&2\\
P16&3&\textbf{2}&1&1&3&2\\
P17&\textbf{1}&\textbf{1}&\textbf{1}&\textbf{1}&\textbf{1}&1\\
P18&\textbf{1}&\textbf{1}&\textbf{1}&\textbf{1}&\textbf{1}&1\\
P19&1&1&1&1&1&2\\
P20&\textbf{1}&\textbf{1}&\textbf{1}&\textbf{1}&\textbf{1}&1\\
P21&\textbf{1}&\textbf{1}&\textbf{1}&\textbf{1}&\textbf{1}&1\\
\hline
\multicolumn{1}{l}{\textbf{Correct count}}&14&\textbf{17}&14&15&14\\
\multicolumn{1}{l}{\textbf{Accuracy}}&66.7\%&\textbf{81.0\%}&66.7\%&71.4\%&66.7\%\\
\bottomrule
\end{tabular}
\caption{Result of the preliminary experiment to select the evaluation model for the main experiment, tested with the dataset shown in Table~\ref{tab_preexp}. Boldface in the option numbers shows correct evaluations. Boldface in the correct count and accuracy show the best case.}
\label{tab_preexp_result}
\end{table*}

While this experiment might not be not sufficient to fully characterize the accuracy of the evaluation method, it suggests that the approach is capable of capturing general trends for each combination of regularity identifier LLM and number series category.

\section{Examples of outputs}\label{sec_examples}
Successful examples of valid regularity identification across all eight categories are shown in Table~\ref{tab_result1}. These examples were selected from the descriptions evaluated as valid, representing typical cases that clearly align with human-confirmed results.  
In the arithmetic, geometric, and difference series categories, the expected mathematical regularities are successfully identified. In the quasi-ordered categories, the $+1$ or $-1$ errors in comparison to the simple ordered series were correctly recognized, and the regularities were described while accounting for these errors. In the random-increasing and random cases, although no clear patterns are apparent, the regularity descriptions successfully explain all the values in the series.  
As mentioned in the main text, we do not require the rule descriptions to be capable of reproducing all numbers in the series. Instead, a description is considered valid as long as it does not contradict the given values. Thus, the descriptions are deemed valid.
\begin{table*}[h!]
\centering
\begin{tabular}{p{15mm}p{28mm}p{15mm}p{85mm}}
\toprule
 \textbf{Category} &\textbf{Number series}& \textbf{Model} & \textbf{Result}  \\
\hline
arithmetic & $1, 9, 17, 25, 33,$ ...& Llama3.3&
The series is formed by adding 8 to the previous term, starting from 1, to generate a sequence of numbers that increase by a consistent difference.\\
geometric & $7, -21, 63,-189,$   ~$567,$ ...& GPT-4.1&
Each term in the series is obtained by multiplying the previous term by -3.\\
difference & $9, 10, 15, 24, 37,$ ...& o3&
Beginning with nine, each subsequent term is produced by adding consecutive odd increments that themselves rise by four (1, then 5, then 9, then 13, and so on).\\
quasi-arithmetic & $6, 14, 23, 30, 38,$ ...& o4-mini&
Each term is generated from the previous one by cyclically adding 8, then 9, then 7, and repeating that cycle.\\
quasi-geometric & $1, -5, 25, -125, $ ~ $626,$ ...& Gemini2.5&
The series generally follows a pattern of powers of -5, but the fifth term listed is one greater than this rule suggests.\\
quasi-difference & $2, 11, 20, 32, 44,$ ...& Llama3.3&
The regularity in the series is obtained by adding 9, then 9, then 12, and then 12, and so on, increasing the increment by 3 after every two steps.\\
random-increasing & $20, 30, 33, 41, 51,$ ...& GPT-4.1&
Each term in the series alternately increases by 10 and then by 3 or 8 in succession.\\
random & $86, 24, 59, 48, 32,$ ...& o3&
The sequence consists of two intertwined arithmetic progressions: the numbers in the odd positions fall by twenty-seven each time, whereas those in the even positions rise by twenty-four on every move.\\
\bottomrule
\end{tabular}
\caption{Examples of successful regularity descriptions, confirmed by human evaluation, selected from cases deemed valid.}
\label{tab_result1}
\end{table*}

Examples of invalid regularity descriptions are listed in Table~\ref{tab_result2}. We did not include cases from the arithmetic series, as all instances were correctly identified.  
In the geometric series, the common ratio is a fixed factor, not an increasing one.
For the difference series, the change from 14 to 23 is 9, which is not a prime number.
The explanations for the quasi-arithmetic, quasi-geometric, and quasi-difference series correspond to the pure forms of each series type, without accounting for deviations.
In the random-increasing explanation, only the first three terms are addressed, and the explanation is inconsistent with the remaining two terms.
The explanation for the random case only accounts for the difference between the first two terms.
\begin{table*}[h!]
\centering
\begin{tabular}{p{15mm}p{28mm}p{15mm}p{85mm}}
\toprule
 \textbf{Category} &\textbf{Number series}& \textbf{Model} & \textbf{Result}  \\
\hline
geometric & $4, 20, 100, 500, $ ~ $2500,$ ...& Llama3.3&
The series progresses by multiplying the previous term by a increasing factor, starting with 5, to generate the next term.\\
difference & $9, 14, 23, 36, 53,$ ...& GPT-4.1&
In this series, each term (starting from the second) is obtained by adding consecutive prime numbers to the previous term.\\
quasi-arithmetic & $1, 9, 17, 25, 32,$ ...& o3&
It is simply an arithmetic progression in which every new element is produced by adding eight to the one before it.\\
quasi-geometric & $8, -16, 33, -64, $ ~ $128,$ ...& o4-mini&
Each term alternates in sign and its absolute value doubles that of the term before it.\\
quasi-difference & $7, 15, 30, 52, 80,$ ...& Gemini2.5&
The difference between consecutive terms increases by seven.\\
random-increasing & $5, 8, 14, 24, 31,$ ...&GPT-4.1&
Each number in the series is formed by alternately adding 3 and then 6 to the previous number.\\
random & $54, 74, 24, 5, 23,$ ...& o3&
Every succeeding term is obtained from the preceding one by reversing its digits and then adding or subtracting 29 alternately.\\
\bottomrule
\end{tabular}

\caption{Examples of unsuccessful regularity descriptions, confirmed by human evaluation, selected from cases deemed invalid.}
\label{tab_result2}
\end{table*}


\end{document}